\pdfoutput=1
\documentclass[10pt,twocolumn,letterpaper]{article}

\usepackage{cvpr}              % To produce the CAMERA-READY version
\usepackage{graphicx}
\usepackage{amsmath}
\usepackage{amssymb}
\usepackage{booktabs}

\usepackage[pagebackref,breaklinks,colorlinks]{hyperref}

\usepackage{bm}
\usepackage{multirow}

\usepackage[capitalize]{cleveref}
\crefname{section}{Sec.}{Secs.}
\Crefname{section}{Section}{Sections}
\Crefname{table}{Table}{Tables}
\crefname{table}{Tab.}{Tabs.}

\def\cvprPaperID{7621} % *** Enter the CVPR Paper ID here
\def\confName{CVPR}
\def\confYear{2023}

\begin{document}
\title{DATE: Domain Adaptive Product Seeker for E-commerce}
\author{
Haoyuan Li\textsuperscript{\rm 1}, 
Hao Jiang\textsuperscript{\rm 2}\thanks{Corresponding author.}, 
Tao Jin\textsuperscript{\rm 1}, 
Mengyan Li\textsuperscript{\rm 2}, 
Yan Chen\textsuperscript{\rm 2}, 
Zhijie Lin\textsuperscript{\rm 1}, 
Yang Zhao\textsuperscript{\rm 1}, 
Zhou Zhao\textsuperscript{\rm 1*} \\
\textsuperscript{\rm 1}Zhejiang University, \textsuperscript{\rm 2}Alibaba Group\\
\tt\small \{lihaoyuan, jint\_zju, linzhijie, awalk, zhaozhou\}@zju.edu.cn\\ 
\tt\small \{aoshu.jh, yian.lmy, cy270543\}@alibaba-inc.com}

% \author{Haoyuan Li\\

% \and


\maketitle

%%%%%%%%% ABSTRACT
\begin{abstract}
Product Retrieval (PR) and Grounding (PG), aiming to seek image and object-level products respectively according to a textual query, have attracted great interest recently for better shopping experience.
Owing to the lack of relevant datasets, we collect two large-scale benchmark datasets from Taobao Mall and Live domains with about 474k and 101k image-query pairs for PR, and manually annotate the object bounding boxes in each image for PG.
As annotating boxes is expensive and time-consuming, we attempt to transfer knowledge from annotated domain to unannotated for PG to achieve un-supervised Domain Adaptation (PG-DA). 
We propose a {\bf D}omain {\bf A}daptive Produc{\bf t} S{\bf e}eker ({\bf DATE}) framework, regarding PR and PG as Product Seeking problem at different levels, to assist the query {\bf date} the product. 
Concretely, we first design a semantics-aggregated feature extractor for each modality to obtain concentrated and comprehensive features for following efficient retrieval and fine-grained grounding tasks. 
Then, we present two cooperative seekers to simultaneously search the image for PR and localize the product for PG. 
Besides, we devise a domain aligner for PG-DA to alleviate uni-modal marginal and multi-modal conditional distribution shift between source and target domains, and design a pseudo box generator to dynamically select reliable instances and generate bounding boxes for further knowledge transfer. 
Extensive experiments show that our DATE achieves satisfactory performance in fully-supervised PR, PG and un-supervised PG-DA. 
Our desensitized datasets will be publicly available here\footnote{\url{https://github.com/Taobao-live/Product-Seeking}}.

\end{abstract}

%%%%%%%%% BODY TEXT
\begin{figure}[t]
\begin{center}
\includegraphics[width=1.0\columnwidth]{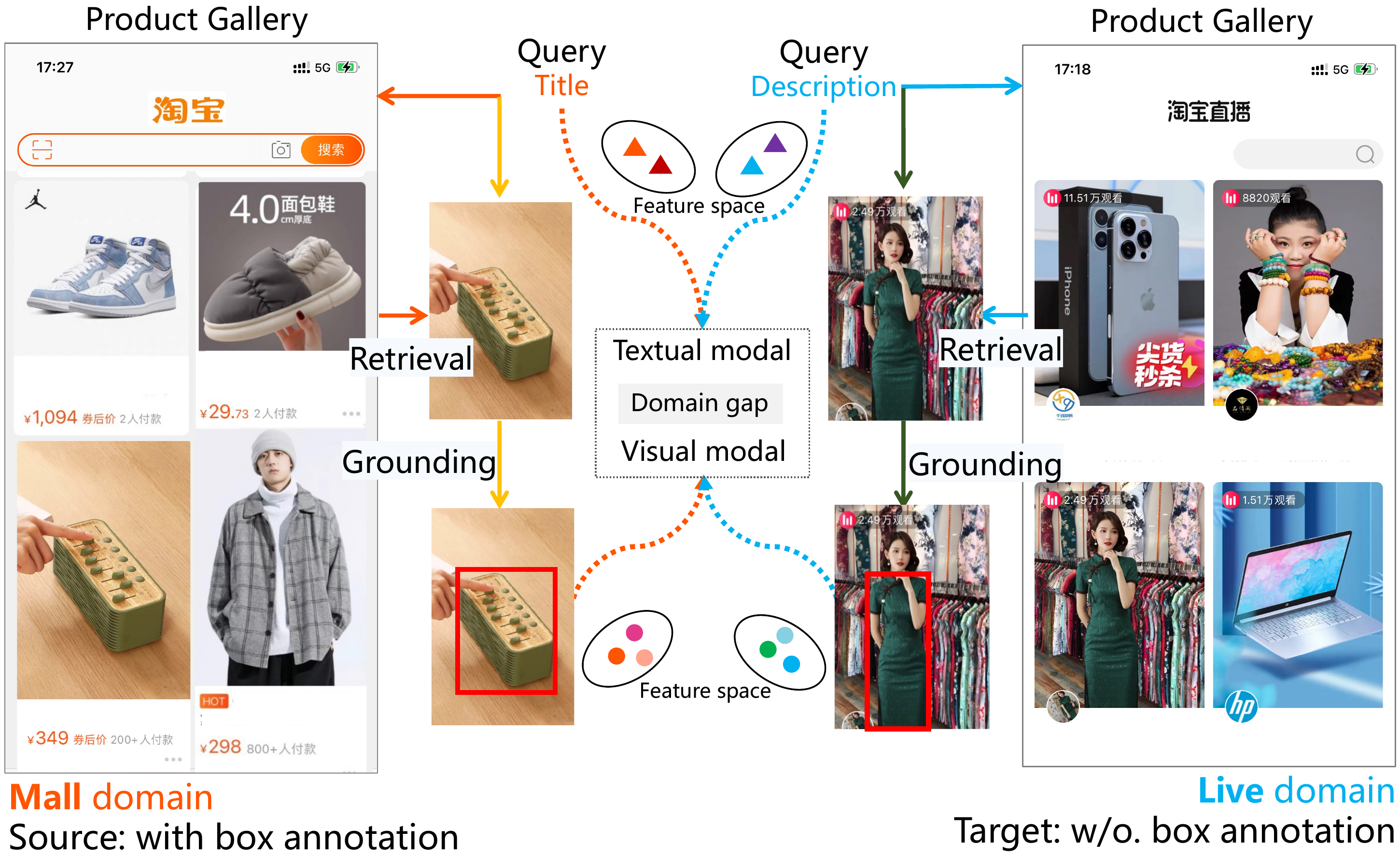}
\end{center}
   \caption{
    Illustration of Product Retrieval (PR) and Grounding (PG) problems on two datasets collected from Taobao Mall and Live. %discrepant domains. 
    (1) Given a text query (i.e. Chinese title or description of a product), PR is to seek the corresponding image-level product from gallery while PG is to seek the object-level product from an image.  %find the image from the gallery and localize the product in the image. 
    (2) We further explore {\bf PG-DA}, which aims to transfer knowledge from the annotated source domain to the unannotated target domain under the influence of multi-modal domain gap to achieve un-supervised PG. %Due to the multi-modal domain gap between two datasets, we 
    % process utilizes annotated source domain data but unannotated target domain data to train models, aiming to transfer knowledge from the source domain to the target. %encounters %requires the model to overcome the influence of domain gap in both visual and textual modalities between two datasets, and
    }
\label{fig:example0}
\end{figure}

\section{Introduction}
Nowadays, with the rapid development of e-commerce and livestreaming, consumers can enjoy shopping on e-mall or various livestreaming platforms. Although the fact that
diverse products can be presented and purchased on screen brings us convenience, we are immersed in this miscellaneous product world. %, and no picnic to search the desired product.  %During a live broadcast, a live streamer would introduce hundreds of products, and the consumer has to find the specific product from the online-shop, which is unfriendly and laboursome.  coarse fine-grained 
Therefore, cross-modal Retrieval \cite{HoangMM17Selective,RevaudICCV19Learning,BrownECCV20Smooth,NgECCV20SOLAR,Zeng20hgr,Bain21Frozen,VisualSparta21Lu} for Product (PR), aiming to seek the corresponding image based on a text query, is significant for boosting holistic product search engine and promoting consumers' shopping experience.
% For instance, since a livestreamer would introduce hundreds of products during a live broadcast, we feel no picnic to capture the desired product immediately.

Besides, provided that the object-level product can be localized on the target product image or live room image according to a query, it will help consumers focus on the desired product and also benefit the downstream vision-to-vision retrieval. And we name this interesting task as Product Grounding (PG) like Visual Grounding \cite{RohrbachECCV16Grounding,Karpathy17Deep,MuAAAI21Disentangled,Liu21Relation_weak,zhaotowards}. %To achieve this goal, we raise a Image-level Product Seeking (I-ProdS) task. However, the corresponding images or live scenes to the query are generally unknown, we further propose the Gallery-level Product Seeking (G-ProdS) to seek for the object-level product from whole dazzling product gallery as Figure 1 shown.
Generally, PR and PG are seen as two separate tasks, but we consider mining the commonalities of PR and PG and regard them as Product Seeking at image-level and object-level respectively. And we design a unified architecture to simultaneously solve PR and PG, which is more time-saving and memory-economical than separate methods.
% However, the existing product search system often fails to return the  satisfactory results, especially in live scenarios. To be specific, it can only list live rooms whose names are associated with query, and our desired product probably is not being introduced due to a livestreamer would introduce hundreds of products during a live broadcast. Thus, it is laboursome to seek the targeted product from dazzling live rooms. Provided the object-level product is precisely localized, consumers are able to swiftly judge whether the product is desired and focus on the targeted one, besides, it benefits the system for better retrieval. 

% To achieve this goal, we raise a novel task Product Seeking (ProdS) with two levels for research, which aims to seek the object-level product based on its title/description query from an image (Image-level ProdS, I-ProdS) or product gallery (Gallery-level ProdS, G-ProdS), as the query dates with product. 
%We divide ProdS task into image-level (I-ProdS) and gallery-level (G-ProdS) according to seeking the product from an image or the gallery.
% in the booming development of e-commerce nowadays, there is no such a special commodity dataset. In addition, with the development of e-commerce business, there are many live streaming products, which are of great research value. Therefore, we collect commodity and live datasets respectively. TO: intro dataset.
% One cannot make bricks without straw, 
To research the PR and PG with great practical application value, we collect two large-scale benchmark Product Seeking datasets TMPS and TLPS from Taobao Mall and Taobao Live domains with about 474k image-title pairs and 101k frame-description pairs respectively, and the locations of object-level products in images are manually annotated. 
% TMPS and TLPS are the first datasets involving e-commerce cross-modal grounding \cite{Hu2017CMN}, and the scale is tens of times larger than existing datasets. 
As annotating bounding box of product is time-consuming and expensive, we explore how to transfer knowledge from an annotated domain to the unannotated one, and achieve un-supervised PG in domain adaptation setting (PG-DA). 
Thus, we propose the {\bf D}omain {\bf A}daptive Produc{\bf t} S{\bf e}eker ({\bf DATE}) to solve the following aspects of the challenging PR, PG and PG-DA problems. 

Firstly, due to the complexity of the mall and live scenarios, discriminative representations of the image and query are prerequisite to accurately localize the object. Considering conventional CNNs are hard to achieve long-distance relation reasoning and full-scale understanding, we utilize and improve the Swin-TF \cite{Liu21Swin} to extract hierarchical and comprehensive features.
As large-scale image seeking is demanding for PR, it is vital to ensure seeking inference is of trivial cost. Thus, we inject [REP] token into Swin-TF to absorb the weighted global semantics, and condense them into a single vector, which will be discriminative and concentrated for following efficient image seeking. And we perform the same semantics-aggregated technique for query feature extraction. 

Secondly, the capacity of both macroscopic image seeking and microcosmic fine-grained object seeking is necessary for PR and PG. %Existing related methods visual retrieval \cite{Zeng20hgr,Bain21Frozen} and visual grounding \cite{Hu2017CMN,Yu2018MAttNet,Liao2020RealTime} and fail to simultaneously handle both tasks, unless cascading the VR and VG, which is time-consuming and memory-wasting.
Therefore, we present two cooperative seekers, where image seeker calculates the cosine similarity between visual and textual concentrated features for PR, and object seeker based on cross-modal interaction transformer directly predicts the coordinates of the product by comprehensive features for PG. We validate the reasonableness of such cooperative strategy through experiments. 

Thirdly, due to the domain gap between two datasets as Figure 1 shown,  applying the model straightway to test on target domain will cause performance degeneration severely for PG-DA. To the best of our knowledge, this is the first work to consider un-supervised Visual Grounding in domain adaptation setting, and  most uni-modal DA  \cite{liu2018cross,long2018conditional,clinchant2016domain} and multi-modal DA \cite{chao2018cross,Chen2021Mind} methods are not directly applicable in our complicated object seeking.
Therefore, we devise a domain aligner based on Maximum Mean Discrepancy to align the domain by minimizing uni-modal marginal distribution and multi-modal conditional distribution divergence between source and target domains, and design a dynamic pseudo bounding box generator to select similar instances in target domain and generate reliable boxes for knowledge transfer.% which achieves satisfactory performance. 

To summarize, the contributions of this paper are as follows:
\begin{itemize} 
\item We collect and manually annotate two large-scale benchmark datasets for PR and PG with great practical application value.
\item We propose a unified framework with semantics-aggregated feature extractor and cooperative seekers to simultaneously solve fully-supervised PR and PG. 
\item We explore un-supervised PG in domain adaptation setting and design the multi-modal domain aligner and dynamic box generator to transfer knowledge.%boost performance. 
\item We conduct extensive experiments which shows that our methods achieve  satisfactory performance in fully-supervised PR, PG and un-supervised PG-DA. 
\end{itemize}
% (1) We collect and manually annotate two large-scale benchmark datasets for PR and PG with great practical application value.
% (2) We propose a unified framework with semantics-aggregated feature extractor and cooperative seekers to simultaneously solve fully-supervised PR and PG effectively. 
% (3) We explore un-supervised PG in domain adaptation setting and design a multi-modal domain aligner and dynamic box generator to boost performance. 
% (4) The experiments show that our DATE achieves competitive performance compared with state-of-the-art Retrieval and Grounding methods in fully-supervised PR and PG, and acquires satisfactory results in un-supervised PG-DA.%, which indicates the effectiveness of our proposed methods.
% \item We conduct extensive experiments to validate 

\begin{figure*}[t]
\begin{center}
\includegraphics[width=2\columnwidth]{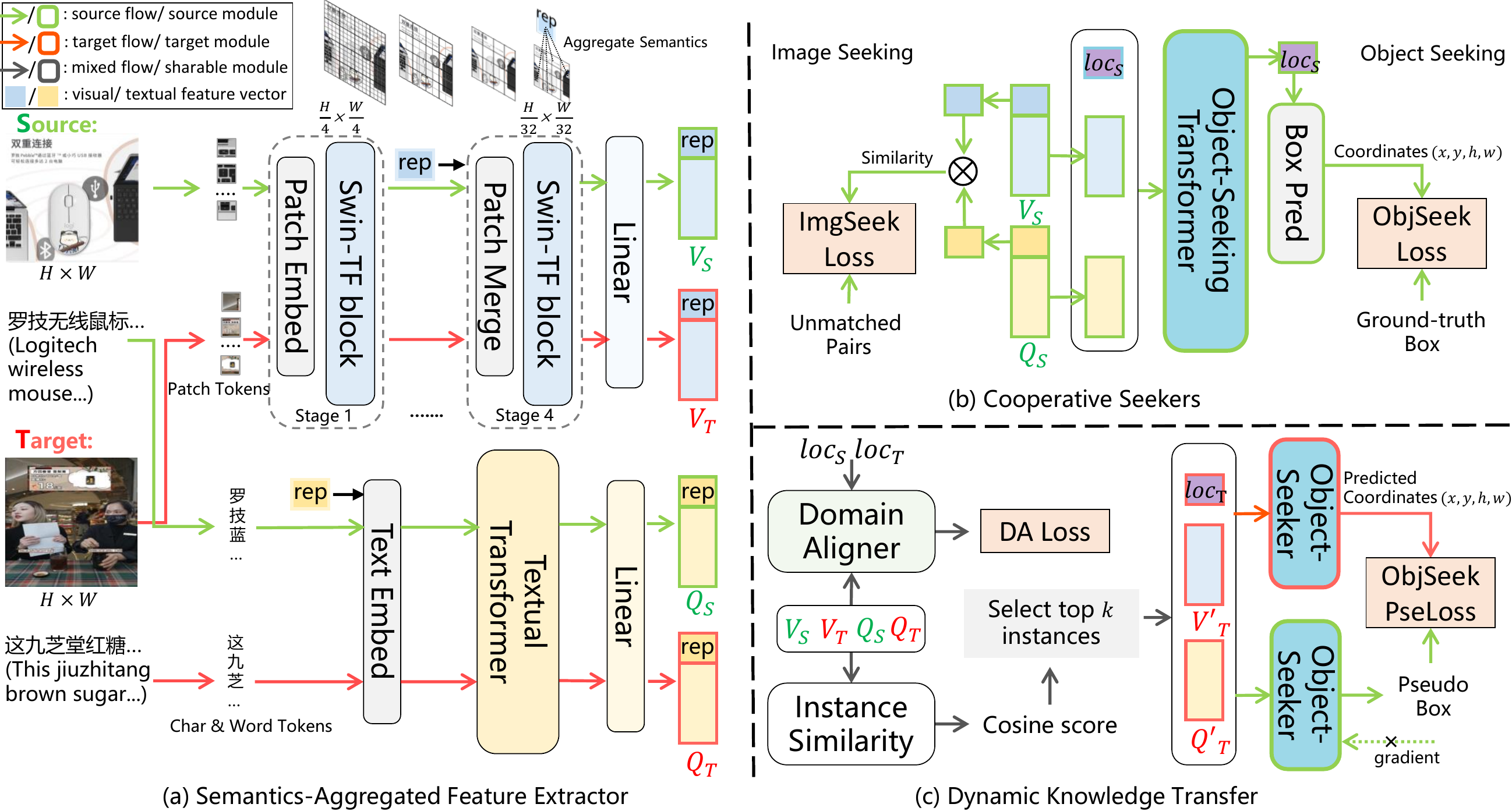}
\end{center}
   \caption{Overview of our DATE. %The green, red and black arrows denote the flow of data from the source, target and mixed domains respectively. %, and only the green flow in Moudle (a) and (b) is performed for fully-supervised ProdS, while whole data flow is for un-supervised ProdS-DA. Concretely, 
   (a) is the feature extractor, applying the semantics-aggregated transformers to obtain image and query features. (b) is the cooperative seekers, calculating the similarity to seek the image for PR and predicting coordinates to seek the object for PG. (c) includes a domain aligner to minimize distribution divergence between source and target domains and a pseudo box generator to select reliable instances and generate bounding boxes for knowledge transfer in PG-DA.
   }
\label{fig:framework}
\end{figure*}

\section{Related Work}
\subsection{Visual Retrieval}
% The tasks most relevant to ProdS are visual retrieval (VR) and grounding (VG). %In previous studies, visual-language retrieval (VLR) and grounding (VLG) were generally considered to be two independent tasks.
Given a text query, Visual Retrieval (VR) \cite{HoangMM17Selective,RevaudICCV19Learning,BrownECCV20Smooth,NgECCV20SOLAR,Zeng20hgr,Bain21Frozen} aims to find the corresponding image/video in a library.
The common latent space based methods \cite{Zeng20hgr,Bain21Frozen} have been proven their effectiveness, which first extract the visual and textual features and map them into a common latent space to directly measure vision-language similarity. 
% Therefore, what matters are the approaches of multi-modal feature extraction and similarity learning. 
Representatively, \cite{Faghri18VSE} applies CNN and RNN to encode images and sentences respectively, and learn image-caption matching based on ranking loss. \cite{Zeng20hgr} proposes a semantic graph to generate multi-level visual embeddings and aggregate results from the hierarchical levels for the overall cross-modal similarity. %decomposes vision-language matching into global-to-local levels
Recently, transformer \cite{Vaswani17TF} exhibits better performance in Natural Language Processing \cite{Devlin18BERT,He2021Deberta}, Computer Vision \cite{Dosovitskiy21ViT,Carion20DETR,jin2022interaction,jin2020sbat,jin2019low} and multi-modal area \cite{yin2022mlslt,yin2021simulslt,GASLT,huang2022prodiff,huanggenerspeech,jin2020dual,lin2021simullr,xia2022video} than previous architecture, especially for global information understanding.  % can scan through the input parallelly and aggregates the information of the whole sequence with adaptive weights.  Compared to RNNs and CNNs, 
Unsuprisingly, there is an increasing effort on repurposing such powerful models \cite{Gabeur20Multi,Zhu2020ActBERT,Bain21Frozen,Kim2021ViLT} for VR. They apply transformer to learn joint multi-mmodal representations and model detailed cross-modal relation, which achieves satisfactory performance.

\subsection{Visual Grounding}
The paradigm of Visual Grounding (VG) \cite{RohrbachECCV16Grounding,Karpathy17Deep,MuAAAI21Disentangled,Liu21Relation_weak}, which aims to localize the objects on an image, is similar as Visual Retrieval (VR), they are both to search the best matching part in visual signals according to the text query. Compared to VR, modeling fine-grained internal relations of the image is more significant for VG. In early work, two-stage methods \cite{Chen2017Query,Hu2017CMN,Yu2018MAttNet} were widely used, which first generate candidate object proposals, then leverage the language descriptions to select the most relevant object, by leveraging off-the-shelf detectors or proposal generators to ensure recall. 
However, the computation-intensive proposal generation is time-consuming and also limits the performance of these methods, one-stage methods \cite{YangICCV19FAOA,Liao2020RealTime} concentrate on localizing the referred object directly. Concretely, \cite{YangICCV19FAOA} fuses the linguistic feature into visual feature maps and predict bounding box directly in a sliding-window manner.
Recently, \cite{Deng2021TransVG} re-formulates VG as a coordinates regression problem and applies transformer to solve it. %Therefore, based on Transformer, we can design different prediction heads to simultaneously handle various problems. 
% However, existing CMR and CMG methods fail to solve our ProdS, due to the capacity of both image-level and object-level seekings is required for ProdS.

Generally, VR and VG are regarded as two separate problems. In this paper, we mine the commonalities of the two problems and design a unified architecture based on cooperative seeking to efficiently solve VR and VG effectively. 

\subsection{Un-supervised Domain Adaptation} 
% UDA \cite{Zhang2020Clarinet}
% Domain Adaptation (DA) aims to transfer the knowledge from the label-rich source domain to the target domain. % Generally, domain adaptation can be divided into semi-supervised domain adaptation by using both the labeled and unlabeled data in the target domain, and unsupervised domain adaptation by only using unlabeled data in the target domain 
Unsupervised domain adaptation (UDA) aims to transfer knowledge from the annotated source domain to the unlabelled target domain, and the challenge is how to overcome the influence of domain gap. In uni-modal tasks applications, several UDA techniques have been explored, including aligning the cross-domain feature distribution \cite{gretton2006kernel,liu2018cross}, applying adversarial learning strategy  \cite{Bousmalis2016Domain,long2018conditional} or reconstruction method \cite{clinchant2016domain} to obtain domain-invariant features. And \cite{courty2016optimal} uses optimal transport to estimate the discrepancy between the two distributions and exploits labels from the source domain. %These methods either rely on finite, predefined and discrete categorical labels or focus on uni-modal retrieval problem.%However, these approaches cannot directly apply to our un-supervised PG since most UDA methods only tackle uni-modal situations, while multi-modal features alignment is required in our setting. %Besides, many methods make use of discrete class labels which are consequently not applicable in our complicated ProdS. 
Different from the works described above, our task is cross-modal in nature, which is more challenging due to the heterogeneous gap between different modalities.
In multi-modal area, few works have considered UDA, \cite{chao2018cross} studies the cross-dataset adaptation for visual question answering, \cite{Chen2021Mind} studies the video-text retrieval with pseudo-labelling algorithm. To the best of our knowledge, this is the first work to consider un-supervised Visual Grounding in domain adaptation setting. %investigates·

\begin{figure}[t]
\begin{center}
\setlength{\abovecaptionskip}{10pt}
\includegraphics[width=1\columnwidth]{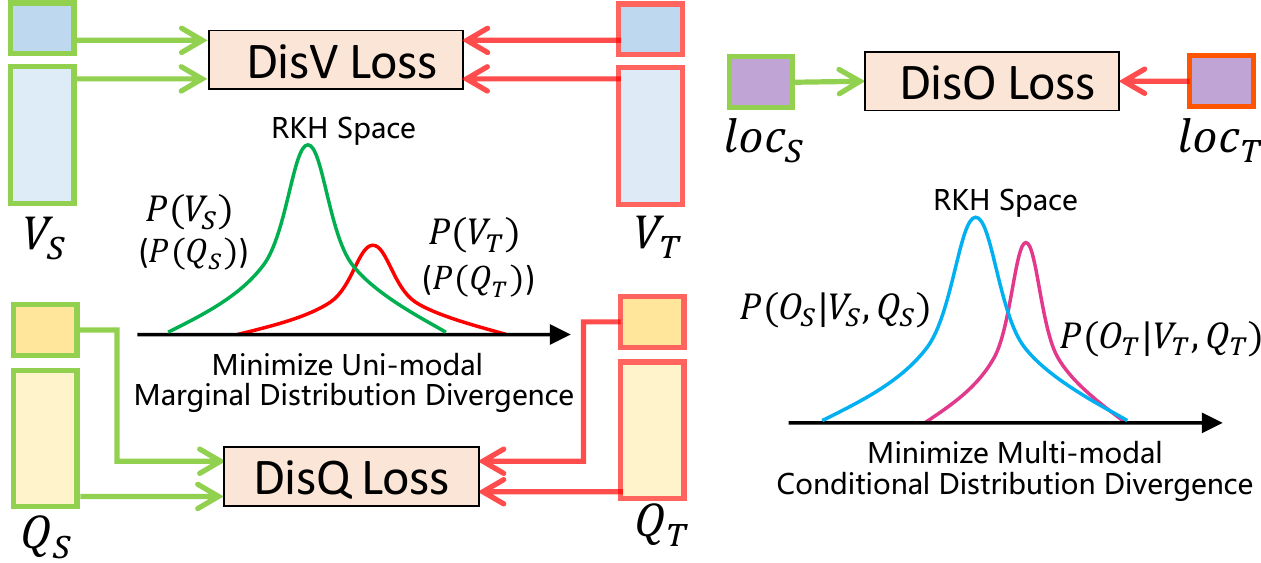}
\end{center}
   \caption{The multi-modal domain aligner. 
   }
\label{fig:framework_da}
\end{figure}

\section{Proposed DATE} 
\subsection{Problem Formulation} 
In this paper, we explore fully-supervised PR and PG, and un-supervised PG-DA in domain adaptation setting. In the next, we will formulate them. 

{\noindent \bf PR and PG.} 
%, note that the former can be regarded as the pre-training of the latter. 
We collect a fully-annotated dataset $\{V,Q,O\}$, given a textual query $Q_i$ in query set $Q$, PR and PG aim to seek the image-level product $V_{Q_i}$ from whole image gallery $V$, and object-level product $O_{Q_i}$ from an matched image $V_{Q_i}$. The $O$ is the bounding box annotation. % Thus, G-ProdS requires to achieve both image and object seekings.
% Note that objects $O=\{o_1,...o_m\}$ are the bounding box annotations. we collect a fully-annotated dataset, which is also as source domain  \boldsymbol{X}^{s}=\left\{\left\{\boldsymbol{V}_{i}^{s}\right\}_{i=1}^{I^{s}},\left\{\boldsymbol{Q}_{j}^{s}\right\}_{j=1}^{J^{s}},\left\{\boldsymbol{B}_{j}^{s}=\left(s_{j}, e_{j}\right)\right\}_{j=1}^{J^{s}}\right\}

{\noindent \bf PG-DA.} 
We have access to a fully-annotated source domain $\mathcal{S}=\left\{ V^S, Q^S, O^S \right\}$, and an unannotated target domain $\mathcal{T}=\left\{V^T, Q^T\right\}$ without box annotation $O^T$. The goal of PG-DA is to transfer the knowledge from $\mathcal{S}$ to $\mathcal{T}$, and %achieve satisfactory performance 
seek the object-level product on $\mathcal{T}$. %In this paper, we only explore the image-level ProdS-DA. 

% \subsection{Overall Architecture}

\subsection{Semantics-Aggregated Feature Extractor}
As Figure \ref{fig:framework}(a) shown, for both settings, we share the feature extractor, which can aggregate the global semantics of each modality for image seeking as well as capture comprehensive and context-aware features for object seeking. %discriminative

{\noindent \bf Image Stream.} Given a RGB image $v$, we first split it into non-overlapping patches, then we refer to Swin-TF \cite{Liu21Swin} for hierarchical feature extraction. Swin is mainly through the stack of patch merging module and Swin Transformer block to achieve 4-stage encoding, and the resolution is halved at each stage to acquire hierarchical features. The original Swin-TF utilizes average pooling to obtain image representation vector, ignoring the difference in importance of each token for semantics extraction. For promotion, we append a learnable [REP] token in front of visual token sequence during 4th stage, which is involved in the computation of self-attention and absorbs the weighted global image features. After the 4th stage, we can acquire the semantics-aggregated visual feature, and we name this advanced visual encoder as SA-Swin.
Next we apply a linear layer to project them into dimension $d$ to obtain $\bm{V}_{SA} = [V_{rep}, \bm{V}] \in {R}^{d \times (1+N_{v})}$, where $N_{v}$ is the number of visual tokens,  $V_{rep}$ and $\bm{V}$ are concentrated and comprehensive features respectively. 

% we first apply a commonly used CNN backbone (e.g ResNet \cite{He2016resnet}) to extract the   low-level and generalized visual feature map ${v_{mp}} \in {R}^{c \times  × h_{mp} \times w_{mp}}$. Then, we reduce the channel dimension $c$ to cross-modal common space dimension $d$ by a 1×1 convolutional layer. After that, we flatten the map into sequential tokens $T_v \in {R}^{d \times N_{v}}$, where $N_{v}=h_{mp} \times w_{mp}$. 
% To aggregate the global semantics of the image into a single vector, we append a [REP] token at the beginning position of the visual tokens as
% \begin{equation}
%     T^{\prime}_v = [T^{rep}_v, T^1_v, ..., T^{N_v}_v]
% \end{equation}
% to obtain tokens $T^{\prime}_v \in {R}^{d \times (1+N_{v})}$. 
% Further, to make the visual transformer sensitive to the original 2D positions of input tokens, utilize sine spatial position encodings as the supplementary of visual feature. 
% To further extract high-level semantics, we exploit a visual transformer which is composed of a stack of 6 standard transformer encoder layers and acquire semantics-aggregated and context-aware visual features $\bm{f}^{SA}_v = [f^{rep}_v, \bm{f}_v] \in {R}^{d \times (1+N_{v})}$. 
% Each transformer encoder layer includes a multi-head self-attention layer and an FFN. There are 8 heads in the multi-head attention layer, and 2 FC layers followed by ReLU activation layers in the FFN.% The output channel dimensions of these 2 FC layers are 2048 and 256, respectively. 

{\noindent \bf Query Stream.} 
% In linguistic stream, we includes a token embedding layer and a linguistic transformer
Given a textual query $q$, we first split it into character-level sequence and convert each character into a one-hot vector. After that, we tokenize each one-hot vector into a dense language vector in the embedding layer.  Similar to image stream, we append a [REP] token in front of the tokenized query sequence to aggregate the global semantics. Note that the visual and textual [REP] tokens are independent for respective aggregation. 
Next we take all tokens into a textual transformer to produce the semantics-aggregated query features. Then we project them into the common space dimension $d$ as image stream, to obtain $\bm{Q}_{SA} = [Q_{rep}, \bm{Q}] \in {R}^{d \times (1+N_{q})}$, where $N_{q}$ is the number of textual tokens. 

\subsection{Cooperative Seekers}
After acquiring common space image feature $\bm{V}_{SA} = [V_{rep}, \bm{V}]$ and query feature $\bm{Q}_{SA} = [Q_{rep}, \bm{Q}]$, as Figure \ref{fig:framework}(b) shown, we design two cooperative seekers to search the matched image and localize the object on this image. Next we describe the responsibility of our two seekers.

{\noindent  \bf Image Seekers for PR.}
The goal of the image seeker is to search the image corresponds to a query. we can directly compute the cosine distance between concentrated features $V_{rep}$ and $Q_{rep}$ to measure the simliarity between image and query, which is time-efficient to search the most similar item and ensures seeking inference is of trivial cost. 
Given a batch $\mathcal{B}$ with $B$ image-text pairs during training, we calculate the  text-to-vision similarity as
\begin{equation}
    p^{q2v}(q) =\frac{\exp (l\cdot s(V_{rep}, Q_{rep})\cdot m^{q2v}) }{\sum_{v \in \mathcal{B}} \exp (l\cdot s(V_{rep}, Q_{rep})\cdot m^{q2v}) } 
\end{equation}
\begin{equation}
    {m^{q2v}}=\frac{\exp \left( { \tau } \cdot {s}\left(V_{rep}, Q_{rep} \right)\right)}
    {\sum_{q \in \mathcal{B}} \exp \left(\tau \cdot {s}\left(V_{rep}, Q_{rep}\right)\right)}
\end{equation} 
where $p^{q2v}(q)$ is text-to-vision probability distribution, $l$ is a learnable logit scaling parameter, $s(\cdot, \cdot)$ denotes cosine similarity, $m$ denotes the prior matrix to refine the similarity distribution following \cite{cheng2021CAMoE}, $\tau$ represents a temperature hyperparameter. For product retrieval on our datasets, the query (title or description of the product) can be also retrieved by the image, and the vision-to-text similarity is $p^{v2q}(v)$.
% \begin{equation}
%     p^{v2q}(v) &=\frac{\exp (l\cdot s(V_{rep}, Q_{rep})\cdot Pri^{v2q}) }{\sum_{q \in \mathcal{B}} \exp (l\cdot s(V_{rep}, Q_{rep})\cdot Pri^{v2q}) } 
% \end{equation}
% \begin{equation}
%     {Pri^{v2q}}=\frac{\exp \left( { \tau } \cdot {s}\left(V_{rep}, Q_{rep} \right)\right)}
%     {\sum_{v \in \mathcal{B}} \exp \left(\tau \cdot {s}\left(V_{rep}, Q_{rep}\right)\right)}
% \end{equation}
Then, we treat matching pairs in the batch as positives, and all other pairwise combinations are treated as negatives, thus the image seeking loss can act as
\begin{equation}
\begin{aligned}
    % \mathcal{L}_{v2t}&=-\frac{1}{B} \sum_{i}^{B} \log \frac{\exp \left(x_{i}^{\top} y_{i} / \sigma\right)}{\sum_{j=1}^{B} \exp \left(x_{i}^{\top} y_{j} / \sigma\right)} \\
    % L_{t}^{v 2 t}&=-\frac{1}{B} \sum_{i}^{B} \log \frac{\exp \left(l \cdot \operatorname{s}\left(v_{i}, s_{i}\right) \cdot P r_{i, i}^{v 2 t}\right)}{\sum_{j=1}^{B} \exp \left(l \cdot \operatorname{s}\left(v_{i}, s_{j}\right) \cdot \operatorname{Pr}_{i, j}^{v 2 t}\right)} \\
    \mathcal{L}_{ImgS} =\frac{1}{2}\mathbb{E}_{v,q \sim \mathcal{B}}[&H\left({p}^{q2v}(q), {y}^{q2v}(q)\right)  \\
    + &H({p}^{v2q}(v), {y}^{v2q}(v)) ],
\end{aligned}
\end{equation}
where $H(\cdot, \cdot)$  is the cross-entropy formulation, ${y}(\cdot)$  is the ground-truth binary label that positive and negative pairs are 1 and 0 respectively. 

{\noindent \bf Object Seeker for PG.} Different from the image seeker, the ambition of object seeker is to localize the microscopic object-level product on an image, and more sufficient image-query interaction and fine-grained seeking are required. Thus, we leverage comprehensive image and query features $\bm{V}$ and $\bm{Q}$ for object seeking.
We consider apply a transformer to fuse cross-modal tokens adequately, in order to learn how to localize the product during interaction, we frist append a learnable [LOC] token with visual and textual features as $\bm{T}_{O} = [T_{loc}, \bm V, \bm Q] \in {R}^{d \times (1+N_{v}+N_{q})}$.
Then we apply a cross-modal object-seeking transformer to embed $\bm{T}_{O}$ into a common space by performing intra- and inter-modality semantic interaction. Besides, we add learnable modal-type embedding and position embedding to the input of each transformer encoder layer.
% Thanks to the attention mechanism, the correspondence can be freely established between each pair of tokens from the joint entities, regardless of their modality. For example, a visual token can attend to a visual token, and it can also freely attend to a linguistic token. 

% The output state of the [LOC] token $f_loc$ develops a consolidated representation enriched by both visual and textual semantics, and is further leveraged for box coordinates prediction. 
We leverage the output state of the [LOC] token $f_{loc}$ from the object-seeking transformer and attach a regression module to it to predict 4-dim box coordinates. %The regression block is implemented by an MLP to directly obtain the 4-dim box coordinates of the box to be grounded. 
%This simplifies the process of  target assignment and positive/negative examples mining at  the training stage, but it also involves the scale problem. Specifically, the widely used smooth L1 loss tends to be a  large number when we try to predict a large box, while tends  to be small when we try to predict a small one, even if their  predictions have similar relative errors. To address this problem, 
Further, to eliminate the influence of scale problem, we normalize the coordinates of the ground-truth box by the scale of the image and perform the object seeking loss as 
\begin{equation} 
    \mathcal{L}_{ObjS}=\|b-\hat{b}\|_{1}+ G(b, \hat{b}),
    \label{Loss_objs}
\end{equation}
where $G(\cdot,\cdot)$ is GIoU Loss\cite{rezatofighi2019GIOU}, $b=(x, y, w, h)$ and $\hat{b}=(\hat{x}, \hat{y}, \hat{w}, \hat{h})$ are our prediction the normalized ground-truth box respectively. 

So far, PR and PG can be solved simultaneously by the cooperation of two seekers, and our cooperative seeking loss is
\begin{equation}
    \mathcal{L}_{coop} =  \lambda_{co}\mathcal{L}_{ImgS} +  \mathcal{L}_{ObjS}, 
\end{equation}
where $\lambda_{co} \in \mathbb{R}$ are hyperparameters to weigh two losses.

\subsection{Dynamic Knowledge Transfer}
As Figure \ref{fig:framework}(a) shown, we design a knowledge transfer method for PG-DA, including a domain aligner to alleviate feature distribution shift and a dynamic pseudo box generator to promote transfer.

{\noindent \bf Domain Aligner.} As Sec 3.3, we extract visual feature $\bm{V}^S_{SA} = [V^S_{rep}, \bm{V}^S]$ and textual feature $\bm{Q}^S_{SA} = [Q^S_{rep}, \bm{Q}^S]$ from $\mathcal{S}$ domain, and we acquire $\bm{V}^T_{SA} = [V^T_{rep}, \bm{V}^T]$ and $\bm{Q}^T_{SA} = [Q^T_{rep}, \bm{Q}^T]$ from $\mathcal{T}$ domain in the same way. 
To alleviate the domain discrepancy, we design an alignment approach based on Maximum Mean Discrepancy (MMD), which compares two distributions by embedding each distribution in to Reproducing Kernel Hibert Space (RKHS) $\mathcal{H}$ with a kernel function $\phi$. And we utilize  multiple Gaussian Radial Basis Function kernels as $\phi$. Given two marginal distributions ${P}_{X^S}$ and ${P}_{X^T}$ from uni-modal source and target domain respectively, MMD can be expressed as
\begin{equation}
    \operatorname{MMD}_{uni}({P}_{X^S}, {P}_{X^T})=\left\|\mu_{{P}_{X^S}}-\mu_{{P}_{X^T}}\right\|_{\mathcal{H}}.
\end{equation}
In order to compute the inner product of vectors using the kernel function $\phi$ in RKHS, we square MMD as 
\begin{equation}
\begin{aligned}
    \operatorname{MMD}&^{2}_{uni}({P}_{X^S}, {P}_{X^T}) 
    =\left\|\mu_{{P}_{X^S}}-\mu_{{P}_{X^T}}\right\|_{\mathcal{H}}^2 \\
    % =&\left\|\frac{1}{n_S} \sum_{i=1}^{n_S} P\left(x^S_{i}\right)-\frac{1}{n_T} \sum_{j=1}^{n_T} P\left(x^T_{j}\right)\right\|_{\mathcal{H}}^{2} \\
    =&\left\|\frac{1}{n_S^{2}} \sum_{i=1}^{n_S} \sum_{{i^\prime}=1}^{n_S} \phi\left(x^S_{i}, x^S_{i^\prime}\right)-\frac{2}{n_S n_T} \sum_{i=1}^{n_S} \sum_{j=1}^{n_T} \phi\left(x^S_{i}, x^T_{j}\right) \right.\\ &\left.+\frac{1}{n_T^{2}} \sum_{j=1}^{n_T} \sum_{{j^\prime}=1}^{n_T} \phi\left(x^T_{j},x^T_{j^{\prime}}\right)\right\|_{\mathcal{H}}.
\end{aligned}
\end{equation}
Then, we can minimize the distance between visual feature distributions from different domains through $\operatorname{MMD}^{2}_{uni}$ as 
\begin{equation}
\begin{aligned}
    \mathcal{L}_{DisV}=\sum_{v \in \mathcal{B}} [&\operatorname{MMD}^{2}_{uni}({V}_{rep}^{S}, {V}_{rep}^{T})  \\
    +&\operatorname{MMD}^{2}_{uni}(\mu(\boldsymbol{V}^{S}), \mu(\boldsymbol{V}^{T}))],
\end{aligned}
\end{equation}
where $\mu(\cdot)$ is calculating the mean value of $\boldsymbol{V}$ on token dimension. In the same way, we compute $\mathcal{L}_{DisQ}$ for textual feature. After that, we can obtain domain-invariant features.

In addition to the discrepancy of uni-modal marginal distribution, we compute the multi-modal conditional distribution divergence to adjust the output distribution for better adaptation, and the form of MMD computation becomes
\begin{equation}
     \operatorname{MMD}_{mul}[{P}({Y^S|X^S_V,X^S_Q}), {P}({Y^T|X^T_V,X^T_Q})].
\end{equation}
Concretely, we take out the output of [LOC] token $f^S_{loc}$ and $f^T_{loc}$ in object seeking transformer from two domains and minimize $\operatorname{MMD}^{2}_{mul}$ to reduce distance of output feature distribution from different domains as
\begin{equation}
\begin{aligned}
    \mathcal{L}_{DisO}=\sum_{f_{loc}^{S}, f_{loc}^{T} \in \mathcal{B}} &\operatorname{MMD}^{2}_{mul}({f}_{loc}^{S}, {f}_{loc}^{T}).
\end{aligned}
\end{equation}
The total domain alignment loss function is as follows
\begin{equation}
    \mathcal{L}_{DA}=\lambda_{Dv}\mathcal{L}_{DisV}+\lambda_{Dq}\mathcal{L}_{DisQ}+\mathcal{L}_{DisO},
\end{equation}
where $\lambda_{Dv}, \lambda_{Dq} \in \mathbb{R}$ are hyperparameters to weigh losses. 

{\noindent \bf Dynamic Pseudo Box Generator.} 
To further transfer the knowledge from $\mathcal{S}$ to $\mathcal{T}$, we attempt to generate pseudo bounding boxes by model on $\mathcal{S}$ to train the model on $\mathcal{T}$. However, it is unlikely that all data can be precisely boxed by source model, which may result in dissatisfactory performance. Therefore, the instances from $\mathcal{T}$ which are close to $\mathcal{S}$ are relatively reliable to be selected. For more precise selection, we compute the instance similarity between two datasets rather than batches. Thus, given the datasets $\{V^S,Q^S\}$ and $\{V^T,Q^T\}$, we calculate the cosine score of features encoded by semantics-aggregated extractor for every pair $\{V^S,V^T\}$ and $\{Q^S,Q^T\}$ in each modality to obtain similarity matrixs $M_V$ and $M_Q$, and we add them to $M\in[-1,1]^{N_S \times N_T}$, where $N_S$ and $N_T$ are lengths of source and target datasets respectively. Next, we rank the target instances based on the counts exceed the similarity threshold $\theta$ and select the top $k$ percent high-score instances $\{{V^T} ^\prime,{Q^T}^\prime\}$.
Then, we generate the pseudo box $\widetilde{b^{\prime}}$ by source object seeker and predict the coordinate ${b^{\prime}}$ by target object seeker. Like Eq. \ref{Loss_objs}, we perform the pseudo object seeking loss as 
\begin{equation} 
    \mathcal{L}_{PObjS}=\|b^{\prime}-\widetilde{b^{\prime}}\|_{1}+ G(b^{\prime}, \widetilde{b^{\prime}}).
\end{equation}
We compute $M$ each epoch after executing box generation, and the selected instances are dynamically updated. With the constant knowledge transfer, more instances can be labeled correctly, and hyper-parameter ratio $k$ will be increased.
The total knowledge transfer loss function is as follows
\begin{equation}
\mathcal{L}_{KT}=\mathcal{L}_{DA}+\lambda_{PO}\mathcal{L}_{PObjS},
\end{equation}
where $\lambda_{PO} \in \mathbb{R}$ are hyperparameters to weigh losses. 

\subsection{Training and Testing}
{\noindent  \bf Fully-supervised PR and PG.} 
We perform $\mathcal{L}_{coop}$ for training, and we search the image of product by image-seeker for PR, and directly predict the coordinates of product on the image by object-seeker for PG during testing.

{\noindent  \bf Un-supervised PG-DA.} 
We train the model in three stages. First, we warm up our model under fully-supervised setting on $\mathcal{S}$ domain by $\mathcal{L}_{stage_1}=\mathcal{L}_{ObjS}$. Next, we perform $\mathcal{L}_{stage_2}=\lambda_{O}\mathcal{L}_{ObjS}+\mathcal{L}_{DA}$ on $\mathcal{S}$ and $\mathcal{T}$ to reduce domain gap. Then, we execute dynamic box generateing and add $\mathcal{L}_{PObjS}$ as $\mathcal{L}_{stage_3}=\lambda_{O}\mathcal{L}_{ObjS}+\mathcal{L}_{KT}$ to further transfer the knowledge. We test the model on $\mathcal{T}$ domain in the same approach as PG.

%(Note that we share the encoders for two domains but construct an another target object-seeking transformer for $\mathcal{T}$ domain)

\begin{table}[t]
    \caption{Performance of Product Retrieval (text-to-vision) on our TMPS and TLPS datasets.}
    \centering
    \setlength\tabcolsep{7pt}%调列距
    \begin{tabular}{l|cccc}
    \hline
        \multirow{2}{*}{Method}  & \multicolumn{4}{c}{\bf TMPS}   \\
        & R@1 & R@5 & R@10 & R@50 \\  \hline 
        Random  & 0.00 & 0.04 & 0.09 & 0.43 \\ 
        VSEpp & 10.23 & 29.24 & 34.42 & 69.73 \\
        ViLT  & 14.39 & 38.42 & 50.74 & {\bf 83.23} \\ 
        \bf DATE  & {\bf 16.32} & {\bf 40.54} & {\bf 51.23} & { 82.58}\\ \hline 
        \multicolumn{1}{c}{} & \multicolumn{4}{c}{\bf TLPS}  \\ \hline 
        Random  & 0.03 & 0.14 & 0.23 & 1.59 \\ 
        VSEpp & 3.41 & 15.33 & 29.12 & 43.24 \\
        ViLT  & 5.38 & 19.29 & 35.95 & 57.48 \\ 
        \bf DATE  & {\bf 6.44} & {\bf 21.71} & {\bf 36.32} & {\bf 59.58}\\ \hline 
    \end{tabular}
    \label{tab:per_PR}
\end{table} 

\begin{table}[t]
    \caption{Performance of Product Grounding on our TMPS and TLPS datasets.}
    \centering
    \setlength\tabcolsep{7pt}%调列距
    \begin{tabular}{l|cc|cc}
    \hline
        \multirow{2}{*}{Method} & \multicolumn{2}{c|}{\bf TMPS} & \multicolumn{2}{c}{\bf TLPS} \\
        & mIoU & Pr@1 & mIoU & Pr@1  \\ \hline 
        Random  & 29.51 & 18.22 & 23.91 & 10.09 \\ 
        MAttNet & 80.71 & 85.33 & 62.12 & 73.24 \\
        FAOA & 76.24 & 83.72 & 61.31 & 69.13 \\ 
        TransVG & 84.52 & 89.50 & 67.11 & 77.93 \\
        \bf DATE  & {\bf 86.67} & {\bf 92.12} & {\bf 70.24} & {\bf 81.43}\\ \hline 
    \end{tabular}
    \label{tab:per_PG}
\end{table}

\section{Experiments}
% \footnote{Our datasets and codes can be found in the found in the supplementary}
\subsection{Our Product Seeking Datasets}

% Previous cross-modal retrieval [17, 29, 45] or visual grounding datasets are not  appropriate for retrieval and grounding tasks, due to the former lack of bounding box annotations and text descriptions in the latter are not unique enough for retrieval. 其实COCO和flickr都是能同时做的。。

% Due to the uniqueness of the product title, our commodity dataset is naturally appropriate for both tasks.%为了任务->标数据，而不是标数据->找任务，前者这么说让人觉得研究更有意义和实际价值，后者感觉是为做任务而做任务

% Compared to existing common cross-modal retrieval datasets \cite{Lin2014coco,Xu2016MSRVTT} or visual grounding datasets \cite{Plummer2015Flickr30k,Yu2016refcoco}, our ALibaba Product and Live Grounding Dataset have two features: 1), 2) 
% {\bf Data Collection} 

% \subsection{Statistics of Our Dataset }\label{data_det}
% We collect two large-scale product seeking datasets TMPS and TLPS for research, which are from Taobao Mall image and Taobao Live video respectively. 

We collect two large-scale Product Seeking datasets from Taobao Mall (TMPS) and Taobao Live (TLPS) with about 474k image-title pairs and 101k frame-description pairs respectively. They are first two benchmark e-commerce datasets involving cross-modal grounding. 
For TMPS, each product item corresponds to a single title, three levels of categories and multiple displayed images with the manually annotated bounding box.
For TLPS, we collect frames and descriptions from the livestreamer in live video streams, and annotate the location of described product. Note that the language in our datasets is mainly Chinese.
The basic statistics about our datasets is in Appendix. We can see the categories of our datasets are diverse and the number of images are tens of times larger than existing datasets. 
%We split TMPS and TLPS into training/validation/testing sets with 397424/39067/37770 and 83135/9109/9103 image-text pairs respectively,
After the collection, we split each dataset into training/validation/testing sets in a 8:1:1 ratio, and we make sure each product is isolated within one set. 

% Besides, the examples and statistics about our datasets and the comparison results with other related datasets are listed in {Appendix \ref{data_det}}. 

% We collect the videos, the descriptions, and the associated product aspects to form the datasets. Each recommended video has been labeled as buyer-generated or fan-generated by the platform. These two kinds of data are originally generated by users with different background knowledge and intentions. Buyers often focus on the general appearance, salient characteristics, and emotions while descriptions generated by fans often reflect deep insights and understandings about the products. Therefore, we regard these two kinds of videos as individual datasets since they

% retrival: without bounding box annotation； vg: text is not unique, not appropriate for retrieval task

% Fig. 1 shows a typical example of our dataset. In the following subsections, we first introduce the construction of the dataset, including generating expressions (Sec. 3.1), discovering distractors (Sec. 3.2), and post-processing (Sec. 3.3). We then analyse the statistics of our dataset and formally define the task in Sec. 3.4 and Sec. 3.5. 

\subsection{Evaluation Metrics}
{\noindent \bf Product Grounding.} Following \cite{Chen2017Query}, we measure the performance by mIoU (mean Intersection over Union) and precision (predicted object is true positive if its IoU with ground-truth box is greater than 0.5).

{\noindent \bf Product Retrieval.} %We use the precision@$n$ evaluation, which only calculates the scores of correct pairs among top-$n$ selectd images (i.e. the performance will be 0 if top-$n$ selectd images are all wrong). 
We use standard retrieval metrics (following  \cite{Zhu2020ActBERT,Bain21Frozen}) to evaluate text-to-vision (t2v) retrieval and  vision-to-text (v2t) retrieval. We measure rank-based performance by R@K. %(where  higher is better) and also report Median Rank (MR, lower is better). 

\begin{table}[t]
    \caption{Performance of Product Grounding-DA on our datasets. (L$\rightarrow $M means we transfer the knowledge from TLPS to TMPS. And F, W, U stand for Fully-, Weakly-, Un-supervised respectively.)}
    \centering
    \setlength\tabcolsep{6pt}%调列距
    \begin{tabular}{l|c|cc|cc}
    \hline
        \multirow{2}{*}{Method} & \multirow{2}{*}{Mode} & \multicolumn{2}{c|}{\bf TMPS} & \multicolumn{2}{c}{\bf TLPS} \\
        & & mIoU & Pr@1 & mIoU & Pr@1 \\  \hline 
        Random & - & 29.51 & 18.22 & 23.91 & 10.09 \\ 
        ARN & W & 70.72 & 73.32 & 51.31 & 53.24 \\ 
        MAF & W & 72.52 & 75.09 & 54.82 & 59.04 \\ 
        FAOA & F & 76.24 & 83.72 & 61.31 & 69.13 \\ 
        \bf DATE & F & {\bf 86.67} & {\bf 92.12} & {\bf 70.24} & {\bf 81.43}\\ \hline
        \multicolumn{2}{c}{} & \multicolumn{2}{c}{\bf L$\rightarrow$M} & \multicolumn{2}{c}{\bf M$\rightarrow$L} \\ \hline 
        Source-only  & U & 75.20 & 83.62 & 59.64 & 67.71 \\ 
        MMD-uni  & U & 76.93 & 84.87 & 60.74 & 69.01 \\ 
        Pseudo-label  & U & 77.02 & 86.23 & 62.87 & 71.48 \\ 
        \bf DATE & U & {\bf 79.92} & {\bf 89.35} & {\bf 64.86} & {\bf 74.75}\\ \hline 
    \end{tabular}
    \label{tab:da}
\end{table}

\begin{table*}[t]
    \caption{Ablation study of Product Retrieval and Grounding on TMPS and TLPS datasets.}
    \centering
    \setlength\tabcolsep{5pt}%调列距
    \begin{tabular}{l|cc|cccc|cc|cccc}
    \hline
        & \multicolumn{6}{c|}{\bf TMPS} & \multicolumn{6}{c}{\bf TLPS} \\
        Method & \multicolumn{2}{c|}{Grounding} & \multicolumn{4}{c|}{T2V Retrieval} & \multicolumn{2}{c|}{Grounding} & \multicolumn{4}{c}{T2V Retrieval} \\
        & mIoU & Pr@1 & R@1 & R@5 & R@10 & R@50 &  mIoU & Pr@1 &  R@1 & R@5 & R@10 & R@50 \\ \hline
        \multicolumn{13}{c}{Visual Feature Extractor}  \\ \hline
        ResNet & 80.73 & 84.13 & 10.85 & 29.10 & 40.82&70.52 & 64.12 & 72.25 & 2.91 & 13.82 & 30.94&49.31 \\
        DETR & 82.29 & 87.71 & 12.12 & 33.52 & 44.52& 74.13 & 66.13 & 76.81 & 4.33 & 16.39 & 32.81&54.91 \\
        Swin & 83.11 & 89.19 & 13.21 & 35.54 & 46.12 & 77.59 &67.31 & 78.35 & 5.01 & 18.56 & 34.14 & 56.25 \\
        SA-DETR & 84.21 & 90.03 & 14.81 & 36.84 & 47.21 & 78.23 & 68.62 & 79.11 & 5.43 & 19.39 & 35.81 &57.28\\
        {\bf SA-Swin} (Ours) & \bf86.67 & \bf92.12 & {\bf 16.32} & {\bf 40.54} & {\bf 51.23} & {\bf 82.58} & \bf70.24 & \bf81.43 & {\bf 6.44} & {\bf 21.71} & {\bf 36.32} & {\bf 59.58}\\ \hline

        \multicolumn{13}{c}{Cooperative Seekers}  \\ \hline
        w/o Rep & 83.11 & 89.19 & 13.21 & 35.54 & 46.12 & 76.59 &67.31 & 78.35 & 5.01 & 18.56 & 34.14 & 55.25 \\
        w/o ObjS & 82.25 & 87.59 & 12.85 & 36.12 & 45.24 & 75.23 & 65.82 & 75.47 & 4.93 & 18.39 & 35.33 & 54.12\\ 
        w/o Rep\&ObjS & 80.45 & 85.31 & 11.78 & 31.17 & 43.23 & 72.23 & 63.21 & 71.91 & 4.13 & 16.53 & 31.82 & 51.10 \\ 
        {\bf Full} (Ours) &\bf86.67 & \bf92.12 & {\bf 16.32} & {\bf 40.54} & {\bf 51.23} & {\bf 82.58} & \bf70.24 & \bf81.43 & {\bf 6.44} & {\bf 21.71} & {\bf 36.32} & {\bf 59.58}\\ \hline
    \end{tabular}
    \label{tab:ab_main}
\end{table*}

\subsection{Performance Comparison and Analysis} 
To evaluate the effectiveness of DATE, we compare it with various related methods (More details of our methods are reported in Appendix). For each task, we apply untrained model to predict results as {\it Random} method to perceive the difficulty of tasks.

{\noindent \bf Product Retrieval.}
% Due to G-ProdS fails to be solved directly by existing methods, we consider  baseline as the cascade of visual retrieval and grounding. 
We re-implement these representative cross-modal retrieval methods to compare with our DATE. 
\begin{itemize}
\item [1)]{\it VSEpp} \cite{Faghri18VSE}, a respectively encoding method based on CNN and RNN. %which applies CNN and RNN to encode images and texts respectively, and uses hinge-based triplet ranking loss to learn image-text matching. %a respectively encoding method based on CNN and RNN.
\item [2)] {\it ViLT} \cite{Kim2021ViLT}, a jointly encoding method based on transformer.%which applies transformer to jointly encode images and texts, and computes matching score by head output of transformer.% We load its pretrained model during re-implement. %a jointly encoding method based on transformer.
\end{itemize}

{\noindent \bf Product Grounding.} 
In addition to cross-modal retrieval baselines above, we re-implement these classic visual grounding baselines to compare with our DATE. 
\begin{itemize}
\item [1)] 
{\it MAttNet} \cite{Yu2018MAttNet}, a two-stage model.%, which applies Faster R-CNN to extract objects, and utilizes language-base and visual attention to focus on image components relevant to query. 
\item [2)] {\it FAOA} \cite{YangICCV19FAOA}, a one-stage model.%, which is straightly fusing a text query’s embedding into the YOLO object detector, augmented by spatial features so as to account for spatial mentions in the query.
\item [3)] {\it TransVG} \cite{Deng2021TransVG}, a regression-based model under transformer architecture.%, which is under fully transformer architecture and formulates the visual grounding as a direct coordinates regression problem. % under transformer architecture in a regression paradigm. 
\end{itemize}

The PR and PG results are presented in Table \ref{tab:per_PR} and Table \ref{tab:per_PG} respectively. We can see that (1) the {\it Random} results in both tasks are pretty low, showing our PR and PG are challenging. (2) The proposed DATE outperforms all the baselines by a large margin, indicating the effectiveness of our method for both PR and PG. (3) Although the performance of {\it TransVG} and {\it ViLT} is little behind ours, they are two separate models, and our method under unified architecture is more time-efficient and memory-saving.%it is the complex combination of two heavy models, and our architecture is more time-efficient and memory-saving.

{\noindent \bf Un-supervised Product Grounding-DA.}
To validate the effectiveness of our DATE in DA setting, we further re-implement these typical weakly-supervised VG baselines for comparison.  
\begin{itemize} 
\item [1)] {\noindent \it ARN} \cite{LiuICCV19ARN}, a reconstruction-based model.% which designs the adaptive grounding module to compute the matching score between each proposal and query, and reconstructs the query with a collaborative loss.%and .
\item [2)] {\noindent \it MAF} \cite{Wang20MAF}, a contrast-based model. %which builds a contrast-based model to guide the alignment between visual and textual representations. 

%It builds the cor- respondence between image region proposal and query in an adaptive manner: adaptive grounding and collabora- tive reconstruction. Specifically, we first extract the sub- ject, location and context features to represent the propos- als and the query respectively. Then, we design the adaptive grounding module to compute the matching score between each proposal and query by a hierarchical attention model. Finally, based on attention score and proposal features, we reconstruct the input query with a collaborative loss of language reconstruction loss, adaptive reconstruction loss, and attribute classification loss. This adaptive mechanism helps our model to alleviate the variance of different refer- ring expressions. Experiments on four large-scale datasets show ARN outperforms existing state-of-the-art methods by a large margin. Qualitative results demonstrate that the proposed ARN can better handle the situation where mul- tiple objects of a particular category situated together1
\end{itemize}
For DA setting, we serve these methods as baselines for comparison.
\begin{itemize} 
\item [1)] {\noindent \it Source-only}, which applies the model trained on source domain to straightway test on the target dataset.
\item [2)] {\noindent \it MMD-uni}, which only utilizes MMD loss to minimize the  uni-modal marginal distribution distance for visual and textual feature.
\item [3)] {\noindent \it Pseudo-label}, which trains the model on target domain entirely based on the pseudo box labels generated by the model trained on source domain.
\end{itemize}

The results are presented in Table \ref{tab:da}, and we can distill the following observations: (1) our un-supervised DATE outperforms all weakly-supervised methods and fully-supervised methods {\it FAOA} significantly, demonstrating the knowledge has been transfered to target domain effectively. (2) {\it Source-only} method degenerates the performance severely due to the huge semantic gap between two domains, and {\it MMD-uni} only achieves slight improvement as the cross-domain discrepanciy fails to reduced sufficiently.
(3) {\it Pseudo-label} enhances limited performance since a number of bad instances are incorrectly labeled which misleads the model, while our DATE can dynamically select instances and generate reliable bounding boxes for transfer and boosting performance. 
%our DATE with can reduce the cross-domain discrepanciy, effectively transfer the annotated knowledge and boost performance. 

\subsection{Ablation Study}
In this section, we study the effect of different visual feature extractors, text options and cooperative seeking strategies in Table \ref{tab:ab_main}.% and Table \ref{tab:ab_text}. %we conduct ablative experiments to verify the effectiveness of each component in our proposed framework

{\noindent \bf Visual Feature Extractor.} We compare our {\it SA-Swin} to {\it ResNet}, {\it DETR}, {\it Swin} and {\it SA-DETR} methods, where {\it ResNet}, {\it DETR} and {\it Swin} apply ResNet-50 \cite{He2016resnet}, DETR-50 \cite{Carion20DETR} Swin-base \cite{Liu21Swin} to extract image features respectively, and leverage the average pooled feature for PR and feed the flattened last feature map as tokens into object-seeking transformer for PG. And {\it SA-DETR} executes the same way as the former methods for PG, but injects the semantics-aggregated token from beginning for PR as {\it SA-Swin} performs.
% \begin{itemize}
%     \item [(1)] {\it {\it ResNet}} \cite{He2016resnet}, we use ResNet-50 to encode images and leverage the average pooled feature for PR and feed flattened the last feature map as tokens into object-seeking transformer for PG.
%     \item [(2)] {\it {\it DETR}} \cite{Carion20DETR}, we apply DETR-50 to encode images and in the same manner as {\it ResNet}.
%     \item [(3)] {\it {\it SA-DETR}} , we apply DETR-50 to 
%     \item [(4)] {\it Swin} \cite{Liu21Swin}
% \end{itemize}
From the results in Table \ref{tab:ab_main}, we can find following interesting points: (1) {\it Swin} surpasses {\it ResNet} and {\it DETR}, illustrating better visual features are extracted by hierarchical transformer. (2) {\it SA-DETR} performs better than {\it Swin} which has more powerful feature extraction ability during cooperative training, demonstrating our designed semantics-aggregated encoder can extract concentrated and comprehensive features for following cooperative seeking for both PR and PG.
% Expectedly, {\it SA-Swin} surpasses others, illustrating better visual features are extracted by hierarchical transformer. 
%Specifically, The we achieve substantial improvements over the previous state-of-the-art methods

\begin{figure}[t]
    \setlength{\abovecaptionskip}{10pt}
    \begin{center}
        \includegraphics[width=1.0\columnwidth]{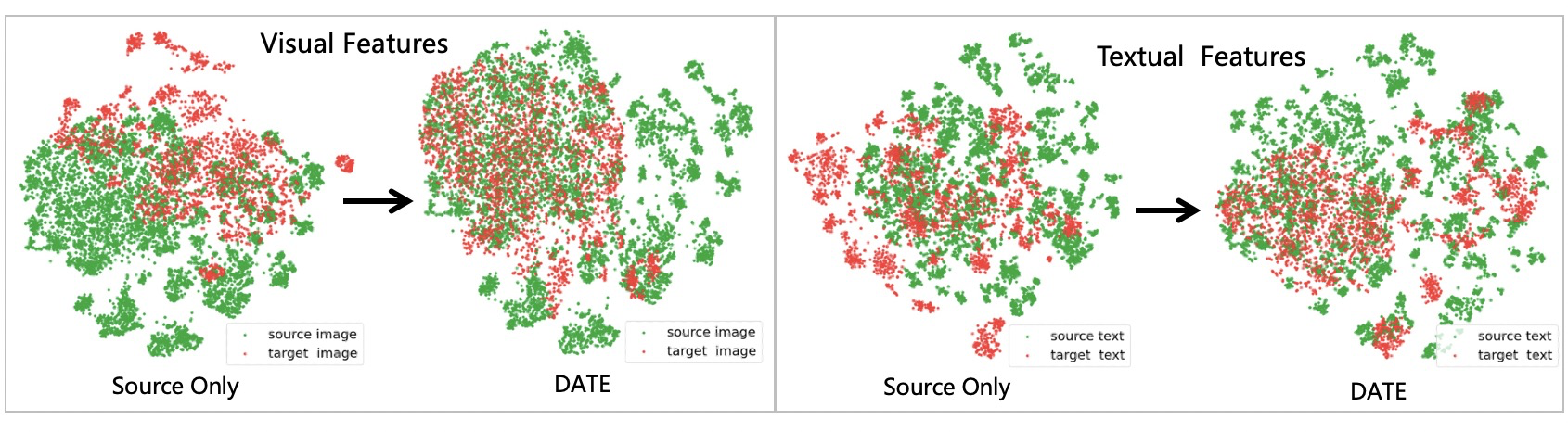}
    \end{center}
    \caption{T-SNE visualization of visual and textual features.}
    \label{fig:viz}
\end{figure}

{\noindent \bf Cooperative Seeking Strategies.} We conduct ablative experiments as follows:  {\bf w/o Rep}: using the average pooling of two modal features for image seeking (PR) rather than [REP] token. {\bf w/o ObjS}: removing object-seeking transformer, and applying an MLP to fuse visual and textual [REP] token for object seeking; {\bf w/o Rep\&ObjS}: using the average pooled feature for both image and object seeking. From Table \ref{tab:ab_main}, we observe that the performance decreases sharply after removing [REP] or ObjS. To analyse: (1) more discriminative representation of image and query can be extracted by weighted vector (i.e. [REP] token) than average pooling, confirming the effectiveness of our semantics-aggregated feature extractor. 
(2) As {\bf w/o Rep} result shown, the performance of object seeking (PG) degenerates although [REP] is not involved in it, which demonstrates such disadvantageous image seeking (PR) approach drags down object seeking (PG) during multi-task learning. 
(3) Image and object levels seeking falls on the shoulder of [REP] tokens in {\bf w/o ObjS} model, which is detrimental for both levels seeking. The above two points prove the reasonableness of our designed cooperative seeking strategy.
% [REP] head can weight every token feature and condense them into a vector 

\begin{figure}[t]
\begin{center}
\includegraphics[width=1.0\columnwidth]{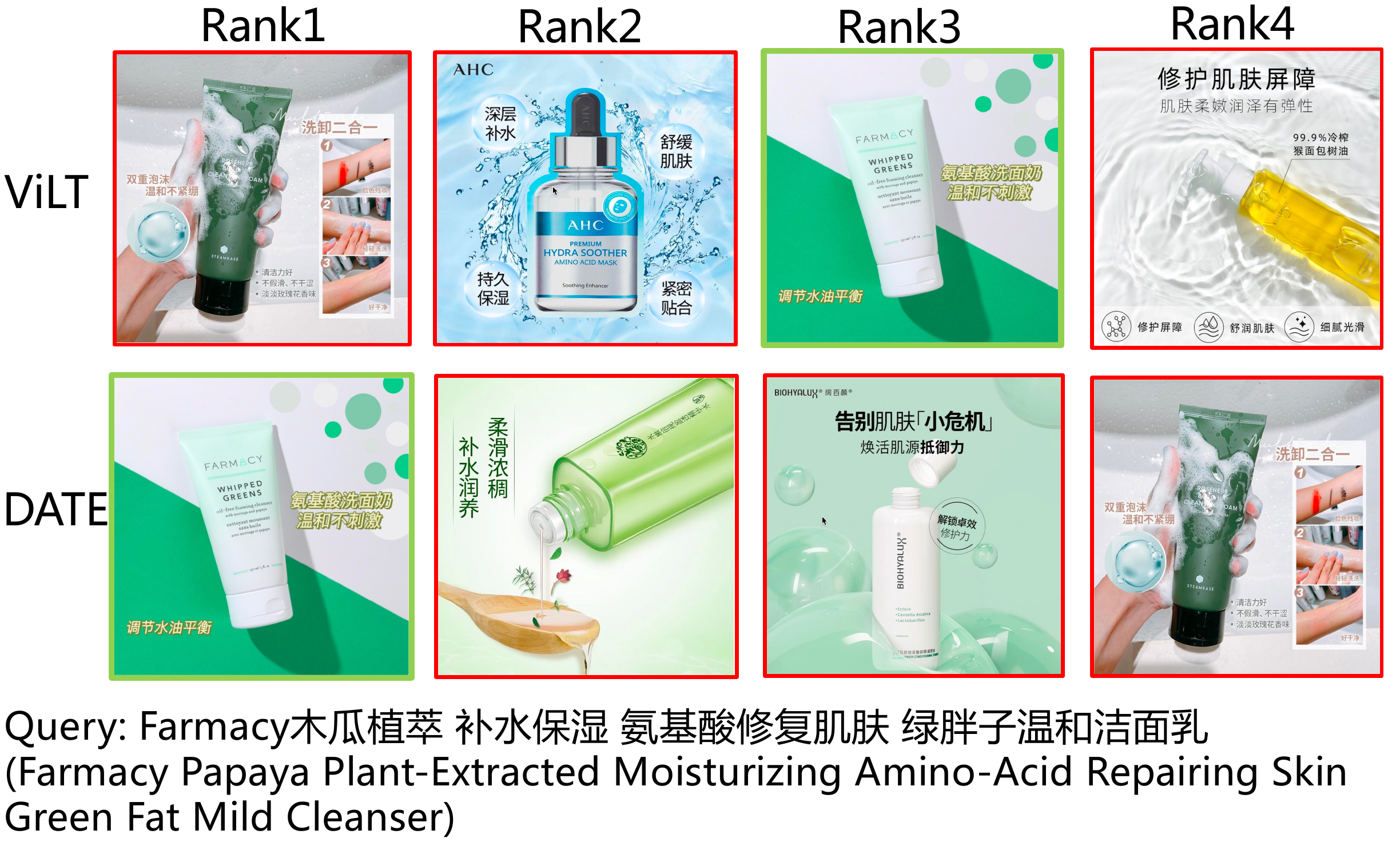}
\end{center}
   \caption{Qualitative results of Product Retrieval sampled from TMPS dataset (green: correct, red: incorrect).}
\label{fig:qua_vr}
\end{figure}

\subsection{Feature Visualization}
To help prove the validity of our DATE, we visualise visual and textual features by T-SNE for TMPS$\rightarrow$TLPS in Figure \ref{fig:viz}, earned by {\it Source-only} baseline and our DATE method. %The cross-domain joint video-text features learnt by our method are clearly clustered tighter than other models, which suggests the effectiveness of aligning the cross-domain video-text distributions, even if without the target domain text queries in advance.
We can observe the shift between source and target domains is apparent, meanwhile there are overlaps in two domains, which is reasonable since a few scenes in Taobao Mall and Live are similar. With our proposed method, the discrepancy in feature distribution of two domains becomes narrow significantly, suggesting our method has effectively aligned two domains. % To a certain extent, it explains why DATE can improve performance. % discriminated

\subsection{Qualitative Analysis}
To qualitatively investigate the effectiveness of our DETA, we compare {\it ViLT} and our DATE for PR as Figure \ref{fig:qua_vr} shown. 
We can find that the image-level product can be sought precisely by our DATE while {\it ViLT} fails to find the correct image until Rank3. Further, the whole top4 results retrieved by DATE are more relevant to the text query than the results from {\it ViLT}, which illustrates the multi-modal semantic understanding and interaction are sufficient through our DATE.
% We can find the following points: (1) The image-level product can be sought precisely by our DATE while {\it ViLT} fails to find the correct image until Rank3. (2) The models can still localize the product on incorrect images, confirming the fine-grained discriminating capacity of object seeker. 
% More examples and qualitative analysis are in Appendix \ref{more_example}.

% \subsection{Online A/B Test}
% We deploy our model DATE on the \emph{Taobao Live Product Retrieval} scenario, which requires the system to retrieve the product livestreamers 
% Originally, the system 
% We conduct online tests on this baseline method and DATE under the framework of the A/B test. We observe that DATE improves the f1-socre by +5.9\% compared to the baseline. These results demonstrate that the object-level product sought by our DATE is accurate and can promote the performance of downstream retrieval.

% \begin{figure}
% \begin{center}
% \fbox{\rule{0pt}{2in} \rule{.9\linewidth}{0pt}}
% \end{center}
%   \caption{Example annotations from our dataset.}
% \label{fig:example}
% \end{figure}

\section{Conclusion}
In this paper, we study the fully-supervised product retrieval (PR) and grounding (PG) and un-supervised PG-DA in domain adaptation setting.
% which aim to seek the image-level and object-level product respectively according to a text query in both fully-supervised setting and un-supervised.
For research, we collect and manually annotate two large-scale benchmark datasets TMPS and TLPS for both PR and PG.
And we propose a DATE framework with the semantics-aggregated feature extractor, efficient cooperative seekers, multi-modal domain aligner and a pseudo bounding box generator to solve the problems effectively on our datasets. We will release the desensitized datasets to promote investigations on product retrieval, product grounding and multi-modal domain adaptation. In the future, we will consider more specific techniques like Optical Character Recognition (OCR) and Human Object Interaction (HOI) to further improve the performance of PR and PG.

\section*{Acknowledgments}
This work was supported by National Natural Science Foundation of China under Grant No.62222211, No.61836002, No.62072397, and a research fund supported by Alibaba.

\newpage

{\small
\bibliographystyle{ieee_fullname}
\bibliography{main.bib}
}

% \appendix
% \clearpage
\end{document}